%% file: 00_MAIN.tex
\documentclass[letterpaper, 10 pt, conference]{class/ieeeconf}  

\IEEEoverridecommandlockouts                              
\usepackage{amsmath}
\usepackage{amssymb}
\usepackage{latexsym}
\usepackage{url}
\usepackage{cite}
\usepackage{relsize}
\usepackage{multirow}
\usepackage{afterpage}
\usepackage{ifthen}
\usepackage{graphicx}
\usepackage{algpseudocode,algorithm,algorithmicx}
\usepackage{subfigure}
\usepackage[keeplastbox]{flushend}
\usepackage{epstopdf}
\usepackage{stackengine}
\usepackage{gensymb}
\usepackage[bookmarks=false, linkcolor=blue, urlcolor=blue, citecolor=blue]{hyperref} 
\usepackage{booktabs}
\usepackage{makecell}
\usepackage{hhline}
\usepackage{lipsum}
\usepackage{mathtools}
\usepackage{color}
\usepackage[10pt]{moresize}

\hypersetup{pdfstartview={FitH}}

\newcommand\red[1]{{\color{red}#1}}

\title{\LARGE \bf Object Captioning and Retrieval with Natural Language}

\author{Anh Nguyen$^{1}$, Thanh-Toan Do$^{2}$, Ian Reid$^{2}$, Darwin G. Caldwell$^{1}$, Nikos G. Tsagarakis$^{1}$%
		\thanks{$^{1}$Anh Nguyen, Darwin G. Caldwell, and Nikos G. Tsagarakis are with the Department of Advanced Robotics, IIT, Italy. {\tt \{Anh.Nguyen, Darwin.Caldwell, Nikos.Tsagarakis\}@iit.it}} 
		\thanks{$^{2}$Thanh-Toan Do and Ian Reid are with the Australian Centre for Robotic Vision (ACRV), University of Adelaide. {\tt \{Thanh-Toan.Do, Ian.Reid\}@adelaide.edu.au}}}

\begin{document}
\input{0_localmacros}

\maketitle
\thispagestyle{empty}
\pagestyle{empty}

\begin{abstract}
We address the problem of jointly learning vision and language to understand the object in a fine-grained manner. The key idea of our approach is the use of object descriptions to provide the detailed understanding of an object. Based on this idea, we propose two new architectures to solve two related problems: object captioning and natural language-based object retrieval. The goal of the object captioning task is to simultaneously detect the object and generate its associated description, while in the object retrieval task, the goal is to localize an object given an input query. We demonstrate that both problems can be solved effectively using hybrid end-to-end CNN-LSTM networks. The experimental results on our new challenging dataset show that our methods outperform recent methods by a fair margin, while providing a detailed understanding of the object and having fast inference time. The source code will be made available.

\end{abstract}

\input{1_intro}

\input{2_rw}

\input{3_acnn}
\input{4_exp}
\input{5_conclusions}

\bibliographystyle{class/IEEEtran}
\bibliography{class/IEEEabrv,class/reference}
   
\end{document}

%% file: 0_localmacros.tex

\newtheorem{problem}{Problem}
\newtheorem{lemma}{Lemma}
\newtheorem{theorem}[lemma]{Theorem}
\newtheorem{claim}{Claim}
\newtheorem{corollary}[lemma]{Corollary}
\newtheorem{definition}[lemma]{Definition}
\newtheorem{proposition}[lemma]{Proposition}
\newtheorem{remark}[lemma]{Remark}
\newenvironment{LabeledProof}[1]{\noindent{\it Proof of #1: }}{\qed}

\def\beq#1\eeq{\begin{equation}#1\end{equation}}
\def\bea#1\eea{\begin{align}#1\end{align}}
\def\beg#1\eeg{\begin{gather}#1\end{gather}}
\def\beqs#1\eeqs{\begin{equation*}#1\end{equation*}}
\def\beas#1\eeas{\begin{align*}#1\end{align*}}
\def\begs#1\eegs{\begin{gather*}#1\end{gather*}}

\newcommand{\poly}{\mathrm{poly}}
\newcommand{\eps}{\epsilon}
\newcommand{\e}{\epsilon}
\newcommand{\polylog}{\mathrm{polylog}}
\newcommand{\rob}[1]{\left( #1 \right)} 
\newcommand{\sqb}[1]{\left[ #1 \right]} 
\newcommand{\cub}[1]{\left\{ #1 \right\} } 
\newcommand{\rb}[1]{\left( #1 \right)} 
\newcommand{\abs}[1]{\left| #1 \right|} 
\newcommand{\zo}{\{0, 1\}}
\newcommand{\zonzo}{\zo^n \to \zo}
\newcommand{\zokzo}{\zo^k \to \zo}
\newcommand{\zot}{\{0,1,2\}}
\newcommand{\en}[1]{\marginpar{\textbf{#1}}}
\newcommand{\efn}[1]{\footnote{\textbf{#1}}}
\newcommand{\vecbm}[1]{\boldmath{#1}} 
\newcommand{\uvec}[1]{\hat{\vec{#1}}}
\newcommand{\thv}{\vecbm{\theta}}
\newcommand{\junk}[1]{}
\newcommand{\var}{\mathop{\mathrm{var}}}
\newcommand{\rank}{\mathop{\mathrm{rank}}}
\newcommand{\diag}{\mathop{\mathrm{diag}}}
\newcommand{\tr}{\mathop{\mathrm{tr}}}
\newcommand{\acos}{\mathop{\mathrm{acos}}}
\newcommand{\atantwo}{\mathop{\mathrm{atan2}}}
\newcommand{\SVD}{\mathop{\mathrm{SVD}}}
\newcommand{\quadf}{\mathop{\mathrm{q}}}
\newcommand{\linterp}{\mathop{\mathrm{l}}}
\newcommand{\sgn}{\mathop{\mathrm{sign}}}
\newcommand{\sym}{\mathop{\mathrm{sym}}}
\newcommand{\avg}{\mathop{\mathrm{avg}}}
\newcommand{\mean}{\mathop{\mathrm{mean}}}
\newcommand{\erf}{\mathop{\mathrm{erf}}}
\newcommand{\grad}{\nabla}
\newcommand{\R}{\mathbb{R}}
\newcommand{\defeq}{\triangleq}
\newcommand{\dims}[2]{[#1\!\times\!#2]}
\newcommand{\sdims}[2]{\mathsmaller{#1\!\times\!#2}}
\newcommand{\udims}[3]{#1}
\newcommand{\udimst}[4]{#1}
\newcommand{\com}[1]{\rhd\text{\emph{#1}}}
\newcommand{\ind}{\hspace{1em}}
\newcommand{\argmin}[1]{\underset{#1}{\operatorname{argmin}}}
\newcommand{\floor}[1]{\left\lfloor{#1}\right\rfloor}
\newcommand{\step}[1]{\vspace{0.5em}\noindent{#1}}
\newcommand{\quat}[1]{\ensuremath{\mathring{\mathbf{#1}}}}
\newcommand{\norm}[1]{\left\lVert#1\right\rVert}
\newcommand{\ignore}[1]{}
\newcommand{\specialcell}[2][c]{\begin{tabular}[#1]{@{}c@{}}#2\end{tabular}}
\newcommand*\Let[2]{\State #1 $\gets$ #2}
\newcommand{\algorithmicbreak}{\textbf{break}}
\newcommand{\Break}{\State \algorithmicbreak}
\newcommand{\ra}[1]{\renewcommand{\arraystretch}{#1}}

\renewcommand{\vec}[1]{\mathbf{#1}} 

\algdef{S}[FOR]{ForEach}[1]{\algorithmicforeach\ #1\ \algorithmicdo}
\algnewcommand\algorithmicforeach{\textbf{for each}}
\algrenewcommand\algorithmicrequire{\textbf{Require:}}
\algrenewcommand\algorithmicensure{\textbf{Ensure:}}
\algnewcommand\algorithmicinput{\textbf{Input:}}
\algnewcommand\INPUT{\item[\algorithmicinput]}
\algnewcommand\algorithmicoutput{\textbf{Output:}}
\algnewcommand\OUTPUT{\item[\algorithmicoutput]}

%% file: 1_intro.tex
\section{INTRODUCTION} \label{Sec:Intro}
Over the past few years, deep learning has become a popular approach to solve visual problems, with traditional problems in computer vision such as image classification~\cite{Alex12}, instance segmentation~\cite{Kaiming17_MaskRCNN_short}, and object detection~\cite{Shaoqing2015} experiencing many mini-revolutions. In spite of these remarkable results, the way these problems are defined prevents them from being widely used in robotic applications. For example, the problem of instance segmentation is formed as a binary segment mask inside a rectangle bounding box. While this is reasonable for computer vision applications, we usually need more information (e.g., object part understanding, grasping frame, etc.) for real-world robotic applications~\cite{AffordanceNet17}.

In this paper, we extend the traditional object detection problem to make it more realistic and suitable for robotic applications. We argue that although recent successful object detectors can achieve reasonable results on a dataset with a few thousand classes~\cite{YOLO9000}, they are still limited by the predefined classes presented during the training. Furthermore, the object detector is also unable to provide more useful information about the object. On the other hand, humans are able to distinguish more than $30,000$ basic categories, and we not only recognize the object based on its category, but also are able to describe the object based on its properties and attributes~\cite{Biederman1987}. Motivated by these limitations, we propose to integrate \textit{natural language} into the object detector. Compared to the traditional object detection approaches that output only the category information, our approach provides a better way to understand the objects by outputting its fine-grained textual description. From this observation, we propose a new method to simultaneously detect the object and generate its caption. Moreover, we show that by using natural language, we can easily adapt an object captioning architecture to a retrieval system, which has excellent potential in many real-world robotic applications~\cite{guadarrama2014}.

\begin{figure}[!t] 
\vspace{0.2cm}
    \centering
 	\includegraphics[width=0.99\linewidth, height=0.75\linewidth]{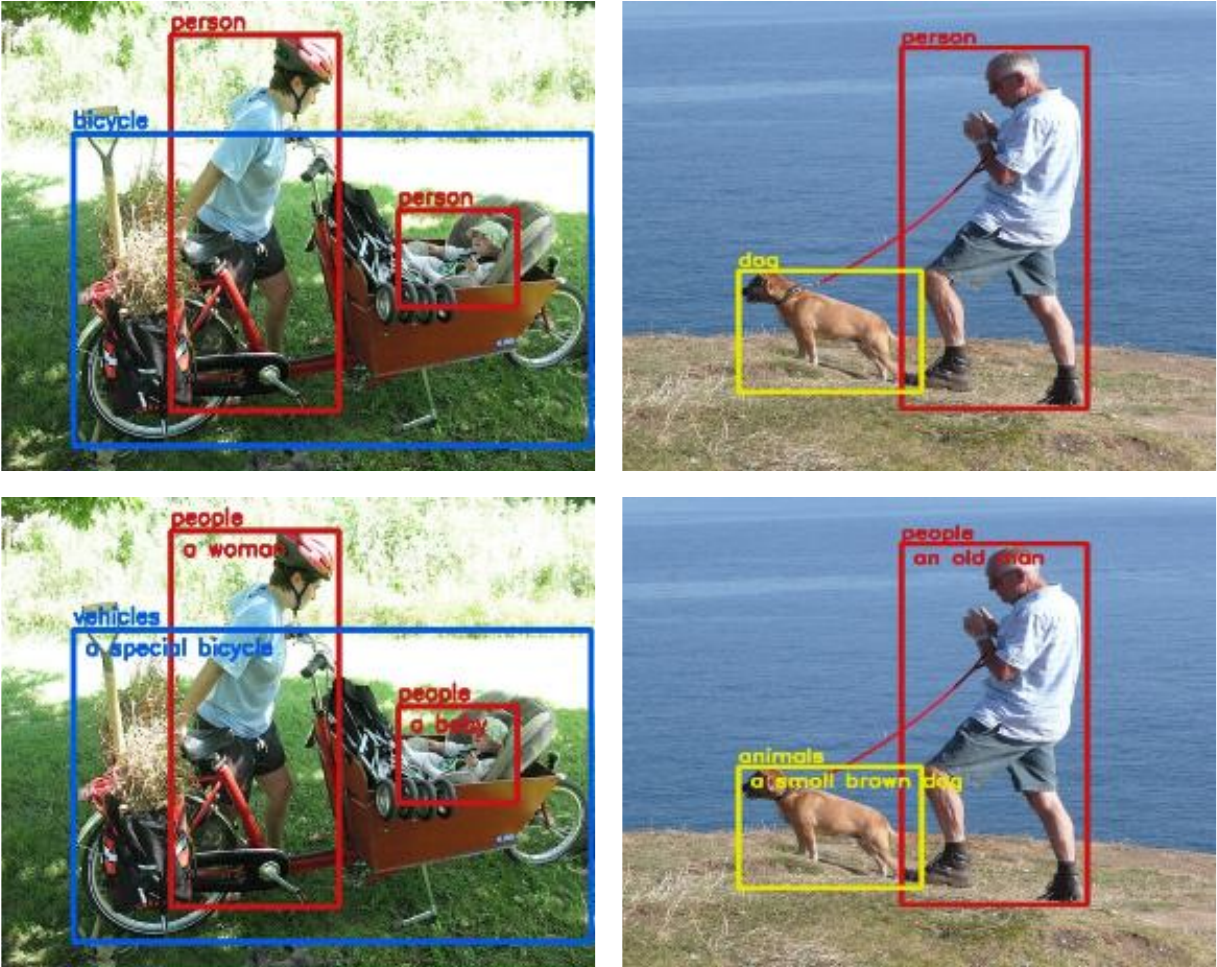} 
    \vspace{1.0ex}
    \caption{A comparison between the traditional object detection problem and our approach. \textbf{Top row}: The groundtruth of the traditional object detection problem which is restricted to the predefined object categories. \textbf{Bottom row}: We propose to use textual caption as the object description, providing a more detailed understanding of each object.}
    \label{Fig:intro} 
    
\end{figure}

In particular, we first define a small set of \textit{superclasses} (e.g., \texttt{animals}, \texttt{vehicles}, etc.), then each object has the caption as its description. This is the main difference between our approach and the traditional object detection problem. The superclass in our usage is a general class, which can contain many (unlimited) fine-grained classes (e.g., the \texttt{animals} class contains all the sub-classes such as \texttt{dog}, \texttt{cat}, \texttt{horse}, etc.). While the superclass provides only the general information, the object descriptions provide the fine-grained understanding of the object (e.g., ``a black dog", ``a little Asian girl", etc.). Fig.~\ref{Fig:intro} shows a comparison between the tradition object detection problem and our approach.

Based on the aforementioned definition, we consider two problems: (1) object captioning and (2) object retrieval using natural language. Our goal in the first problem is to simultaneously detect the object and generate its description. In this way, we have a new object detector that can localize the object while providing more meaningful information. In the second problem, the goal is to localize an object given an input query. This is particularly useful in the human-robot interaction applications since it allows humans to use natural language to communicate with the robot~\cite{guadarrama2014}. We show that both problems can be solved effectively with the deep networks, providing a detailed understanding of the object while achieving the state-of-the-art performance.

The rest of this paper is organized as follows. We first review related work in Section~\ref{Sec:rw}. We then describe two end-to-end architectures for two tasks: object captioning and object retrieval with natural language in Section~\ref{Sec:acnn}. In Section~\ref{Sec:exp}, we present the extensive experimental results on our new challenging dataset. Finally, the conclusion and future work are presented in Section~\ref{Sec:con}.

\junk{
due to its effectiveness~\cite{He2016}.

\red{Compare to the traditional setup which works with closed sets of categories, our approach provides a flexible way to understand the objects, while is not limited by the object categories since any object can be described with natural language.}

     (1) description generation, in which we must generate a text expression that uniquely pinpoints a highlighted object/region in the image and (2) description comprehension, in which we must automati- cally select an object given a text expression that refers to this object (see Figure 1). Most prior work in the litera- ture has focused exclusively on description generation (e.g., [31, 27]). Golland et al. [19] consider generation and com- prehension, but they do not process real world images.

~\cite{Biederman1987} They key idea of our approach is the use of natural language to describe the objects. Compared to the traditional closed sets of categories setup, our approach provide a flexible way to understand the objects, while is not limited by the predefined number of classes since any object can be described with natural language. Our approach is based on the fact that humans . Based on this setting, we propose a new method to simultaneously detect the object and generate its description. Furthermore, we show that by using the natural language, we can easily adapt the detection system to the retrieval system which is potentially useful in many robotic applications.

 the variety of queries when they are used as retrieval systems.
 
  This is still far bellow human ability which is able to distinguish beyond $30,000$ basic categories~\cite{Biederman1987}. Increasing the number of classes, however, will require more manually labeled data, and potentially there may be other theoretical problems such as the imbalance in the number of samples in each class. With the setting based on a closed sets of categories, the solution is limited within this set and is unable to deal with the new class or the variety of queries when they are used as retrieval systems.

Currently, the object detection problem is defined based on a fixed number of classes. TBF.

However, most successful recognition models, to date, are formulated as supervised learning problems, in many cases requiring hundreds, if not thousands, labeled instances to learn a given concept class [10]. This exuberant need for large labeled datasets has limited recognition models to domains with 100’s to few 1000’s of classes. Humans, on the other hand, are able to distinguish beyond 30, 000 basic level categories [5]

Using the retrieval setup de- scribed above, our dense captioning model can also be used to localize arbitrary pieces of text in images. This enables “open-world” object detection, where instead of commit- ting to a fixed set of object classes at training time we can specify object classes using natural language at test-time. We showexample results for this task in Figure 5, where we display the top detections on the test set for several phrases. 

Our model can detect animal parts (“head of a giraffe”, “legs of a zebra”) and also understands some object at- tributes (“red and white sign”, “white tennis shoes”) and in- teractions between objects (“hands holding a phone”). The phrase “front wheel of a bus” is a failure case: the model correctly identifies wheels of buses, but cannot distinguish between the front and back wheel. ~\cite{He2016}

We frame natural language object retrieval as a retrieval task on a set of candidate locations in a given image in this paper, as shown in Figure 1, where candidate locations can come from object proposal methods such as EdgeBox [33]. We observe that simply applying text-based image retrieval systems on the image regions cropped from candidate locations for this task leads to inferior performance, as natural language object retrieval involves spatial configurations of objects and the global scene as context.

 Even when we     recognition models,   In this work, I would like to solve this limitation by extending the traditional object detection problem to the Open-world Object Detection (OOD) problem. The goal is to build an object detector that can detect the super-class and retrieve the most relevant object. For example, in Fig.1-b the object detector should be able to detect super-class such as ``people", ``animal", and also can retrieve the proper object when we ask for ``a little girl", or ``a while dog".

  This lead to the fact that the solutions (even robust one) are limited by the number of object classes, while

   applications based on these problems are limited by the way t  From these traditional problems, new tasks have been formed to address more challenging and realistic scenarios. For example, the problem of instance segmentation is a combination of object detection and binary segmentation, or the problem of affordance segmentation is a combination of object detection and multiclas semantic segmentation. Solving these challenging problems is more difficult, but the results can be used in real-world applications~\cite{•}. However, in practical application scenarios, instead of using a predefined fixed set of object categories, one would often prefer to refer to an object with natural language rather than use a predefined category

   Morbi dolor nulla, malesuada eu, pulvinar at, mollis ac, nulla. Curabitur auctor semper nulla. Donec varius orci eget risus. Duis nibh mi, congue eu, accumsan eleifend, sagittis quis, diam. Duis eget orci sit amet orci dignissim rutrum.

   }

%% file: 2_rw.tex
\section{Related Work} \label{Sec:rw}
In computer vision, detecting objects from RGB images is a well-known task. With the rise of deep learning, recent works design different deep architectures to solve this problem. These architectures can be divided into two groups: region-based~\cite{Shaoqing2015} and non region-based approaches~\cite{YOLO9000}~\cite{Wei2016}. While non region-based methods can achieve real-time performance, region-based architectures provide a more flexible way to adapt the object detection problem to other scenarios~\cite{Kaiming17_MaskRCNN_short}. However, the drawback of the methods in~\cite{Shaoqing2015}~\cite{YOLO9000}~\cite{Wei2016} is their reliance on a fixed set of classes for both training and inference. Therefore, they are unable to deal with a new class or provide more detailed understanding of the object. 

Along with the object detection problem, image captioning is also an active field in computer vision. Current research has shown recurrent neural networks such as LSTM~\cite{Hochreiter97_LSTM} to be effective on this task. Recently, the authors in~\cite{ren2017_deep_RL_caption} proposed to fuse deep reinforcement learning with LSTM and achieved competitive results. While we retain the concept of image captioning for object description, our goal here is more closely related to the dense captioning task~\cite{johnson2016densecap} since we want to generate the caption for each object, not for the entire image. However, unlike~\cite{johnson2016densecap} that generates the captions for all possible regions, our network only generates captions for objects in the superclasses. This allows us to have a more detailed understanding of each object, while still being able to distinguish objects based on their superclasses.

In the robotics community, the work in~\cite{guadarrama2014} introduced a method to localize an object based on a text query. Recently, Hu et al.~\cite{hu2016natural} improved on this by fusing the query text, the spatial, and the global context of the image into three recurrent neural networks. Similarly, the authors in~\cite{Mao2016} introduced a new discriminative training method for this task. Although these methods are able to localize the object based on the input query, their architectures are not end-to-end and unable to run in real-time since the object proposals are generated offline and not trained with the network. With a different viewpoint, Plummer et al.~\cite{Plummer_Flick30K} proposed to localize objects that correspond to the noun phrases of a textual image description. Although our goal in the retrieval task is similar to~\cite{guadarrama2014}~\cite{hu2016natural}~\cite{Mao2016}, the key difference with our approach is the use of an end-to-end architecture, which has a fast inference time and does not depend on any external bounding box extraction method~\cite{hu2016natural}~\cite{Mao2016}.

\junk{

The goal of our work is to integrate the natural language to the object detector. We follow the same concept in~\cite{Shaoqing2015} to build the object detector and introduce two end-to-end architectures to solve two sub-problems: object captioning and object retrieval. Furthermore, based on the output of the network, we demonstrate different robotic applications on a full-size humanoid robot.

also to retrieve the best matched object of an input query as

  The most related task to ours is the phrase localization introduced by Plummer et al. [35], whose goal is to localize objects that corresponds to noun phrases in textual description from an image. TODO: fix
  
which is to divers for robotic application,

Object grounding by natural language. Object grounding by natural language description has recently drawn much attention and several tasks and approaches has been proposed [15, 19, 23, 28, 35]. The most related task to ours is the phrase localization introduced by Plummer et al. [35], whose goal is to localize objects that corresponds to noun phrases in textual description from an image. Many approaches have since been proposed for this task [8, 12, 34, 36, 39, 48, 49]. Chen et al. [8] is the closest to our work in terms of learning region proposals and performing regression conditioned upon a query. Although the tasks of phrase localization and open-vocabulary object detection are similar, most of the phrase localization methods cannot be used for retrieval task due to the reason explained in Sec.

Natural language object retrieval. Based on a bag of words sentence model and embeddings derived from ImageNET classifiers, [10] addresses a similar problem as ours and localizes an object within an image based on a text query. Given a set of candidate object regions, [10] generates text from those candidates represented as bag-of-words using category names predicted from a large scale pretrained classifier and compares the word bags to the query text. Other methods generate visual features from query text and match them to image regions, e.g. through a text-based image search engine [1] or learn a joint embedding of text phrases and visual features. Concurrent with our work, [24] also proposes a recurrent network model to localize objects from given descriptions.

Scene understanding is fundamental problem to enable the robot to interact with the surrounding environment~\cite{}.
The object detection problem is a well-known task in computer vision. Recent approach based on deep learning to solve this problem. Recently, rapid advancements have been made in this field with the rise of deep learning [17]. 

and [14] take an input image and generate a text caption describing it. Recently, methods based on recurrent neural networks [32, 31, 25, 4] have shown to be effective on this task. LRCN [4] is one of these recent successful methods and involves a two-layer LSTM network with the embedded word sequence and image features as input at each time step. We use LRCN as our base network architecture in this work and incorporate spatial configurations and global context into the recurrent model for natural language object retrieval.

}

%% file: 3_acnn.tex
\section{Object Captioning and Retrieval with\\ Natural Language} \label{Sec:acnn}
We start by briefly describing three basic building blocks used in our architecture: Convolutional backbone with Region Proposal Network (RPN) as proposed in Faster R-CNN~\cite{Shaoqing2015}, Long-Short Term Memory (LSTM) network~\cite{Hochreiter97_LSTM} and the embedding of word representations. We then present in details the network architectures for two sub-problems in section~\ref{Sec_acnn_caption} and~\ref{Sec_acnn_retrieval}.

\subsection{Background}

%

\textbf{Convolutional Backbone}
Inspired by~\cite{Kaiming17_MaskRCNN_short}~\cite{AffordanceNet17}, we build an end-to-end network with two branches: the first branch localizes and classifies the object based on its superclass, while the second branch handles the object caption. This architectural methodology was proposed in Faster R-CNN~\cite{Shaoqing2015} and is now widely used since it provides a robust way to extract both the image feature map and the region features. In particular, given an input image, the image feature map is first extracted by a convolutional backbone (e.g., VGG16~\cite{SimonyanZ14}). An RPN that shares the weights with the convolutional backbone is then used to generate candidate bounding boxes (RoIs). For each RoI, unlike Faster R-CNN that uses the RoIPool layer, we use the RoIAlign~\cite{Kaiming17_MaskRCNN_short} layer to robustly pool its corresponding features from the image feature map into a fixed size feature map.

\textbf{LSTM}
In this work, we use LSTM to model the sequential relationship in the object caption. The robustness of the LSTM network is due to the gate mechanism that helps the network encodes the sequential knowledge for long periods of time, while is still remaining sturdy against the vanishing gradient problem. In practice, LSTM is used in many problems~\cite{johnson2016densecap}~\cite{Ramanishka2017cvpr}~\cite{nguyen2017_event}. The LSTM network takes an input $\mathbf{x}_t$ at each time step $t$, and computes the hidden state $\mathbf{h}_t$ and the memory cell state $\mathbf{c}_t$ as follows:

\begin{equation}
\label{Eq_LSTM} 
\begin{aligned} 
{\mathbf{i}_t} &= \sigma ({\mathbf{W}_{xi}}{\mathbf{x}_t} + {\mathbf{W}_{hi}}{\mathbf{h}_{t - 1}} + {\mathbf{b}_i})\\
{\mathbf{f}_t} &= \sigma ({\mathbf{W}_{xf}}{\mathbf{x}_t} + {\mathbf{W}_{hf}}{\mathbf{h}_{t - 1}} + {\mathbf{b}_f})\\
{\mathbf{o}_t} &= \sigma ({\mathbf{W}_{xo}}{\mathbf{x}_t} + {\mathbf{W}_{ho}}{\mathbf{h}_{t - 1}} + {\mathbf{b}_o})\\
{\mathbf{g}_t} &= \phi ({\mathbf{W}_{xg}}{\mathbf{x}_t} + {\mathbf{W}_{hg}}{h_{t - 1}} + {\mathbf{b}_g})\\
{\mathbf{c}_t} &= {\mathbf{f}_t} \odot {\mathbf{c}_{t - 1}} + {\mathbf{i}_t} \odot {\mathbf{g}_t}\\
{\mathbf{h}_t} &= {\mathbf{o}_t} \odot \phi ({\mathbf{c}_t})
\end{aligned}
\end{equation}
where $\odot$ represents element-wise multiplication, $\sigma$ is the sigmod non-linearity, and $\phi$ is the hyperbolic tangent non-linearity function. The weight $\mathbf{W}$ and bias $\mathbf{b}$ are the parameters of the network.

\textbf{Word Embedding} For simplicity, we choose the one-hot encoding technique as our word representation. The one-hot vector $\mathbf{y} \in {\mathbb{R}^{|D|}}$ is a binary vector with only one non-zero entry indicating the index of the current word in the vocabulary. Formally, each value in the one-hot vector $\mathbf{y}$ is defined by:

 \begin{equation}
    \mathbf{y}^j=
    \begin{cases}
      1, & \text{if}\ j=ind(\mathbf{y}) \\
      0, & \text{otherwise}
    \end{cases}
  \end{equation}
where $ind(\mathbf{y})$ is the index of the current word in the dictionary $D$. In practice, we add two extra words to the dictionary (i.e., \textsf{EOC} word to denote the end of the caption, and \textsf{UNK} word to denote the unknown word).

\subsection{Object Captioning}\label{Sec_acnn_caption}

\begin{figure}[!htbp] 
    \centering
	\includegraphics[scale=0.31]{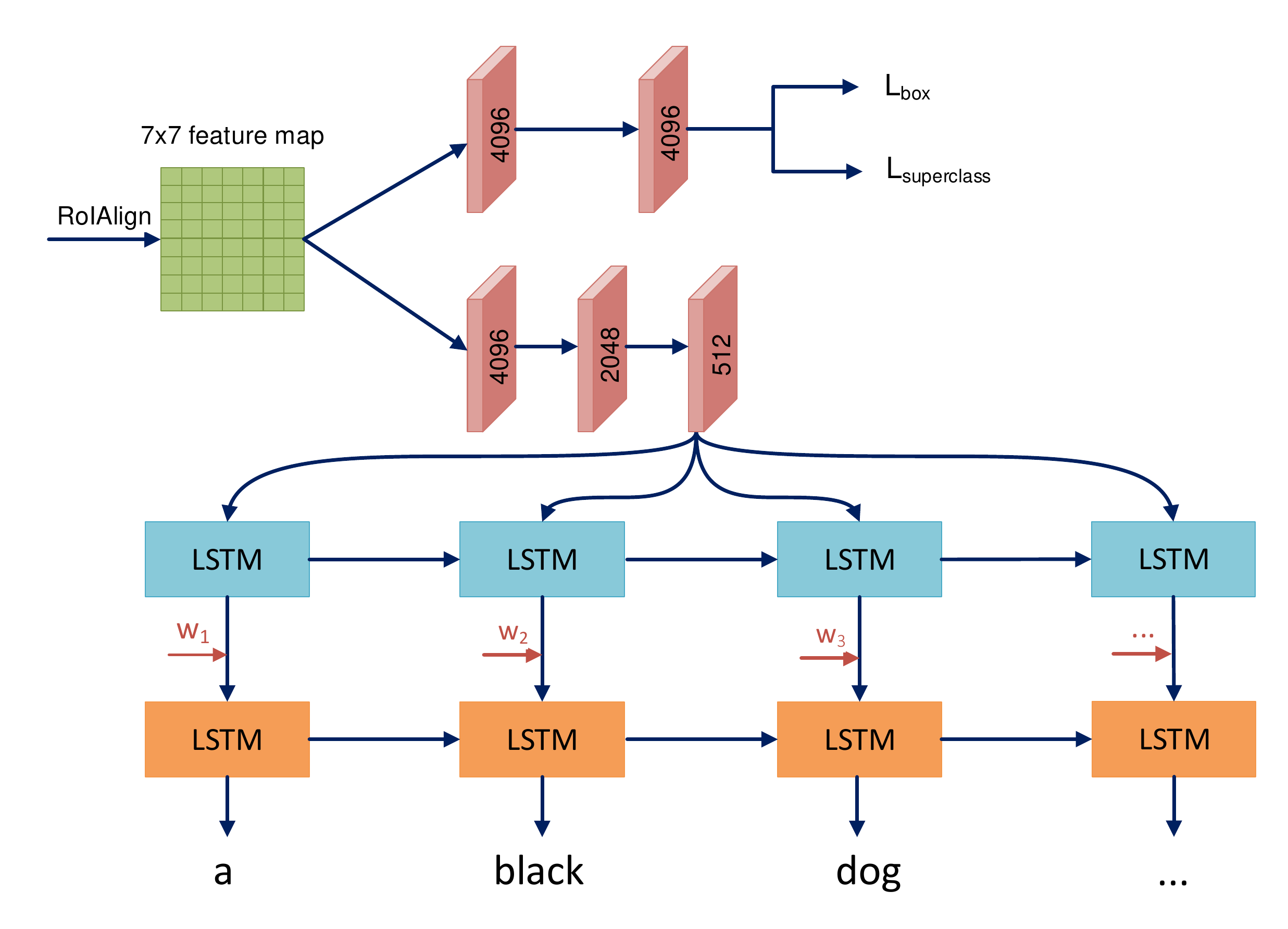} 
    \vspace{1.0ex}
    \caption{An overview of our object captioning network. In the second branch, each positive RoI is first cloned and fed into the $LSTM_{roi}$ layer (in light blue), then the input words and the outputs of $LSTM_{roi}$ layer are combined to become the input for the $LSTM_{word}$ layer (in dark yellow).}
    \label{Fig:overview_caption} 
\end{figure}

In this task, our goal is to simultaneously find the object location, its superclass, and the object caption. While the object location and its superclass are learned using the first branch of the network as in~\cite{Kaiming17_MaskRCNN_short}~\cite{AffordanceNet17}, inspired by~\cite{Ramanishka2017cvpr}~\cite{Nguyen_V2C_ICRA18} we propose to use two LSTM layers to handle the object caption. The first LSTM  layer ($LSTM_{roi}$) encodes the visual information from each RoI provided by the RPN network, while the second layer ($LSTM_{word}$) combines the visual information from the first $LSTM_{roi}$ layer with the input words to generate the object caption. Fig.~\ref{Fig:overview_caption} shows an overview of our object captioning network. 

In particular, we first use the convolutional backbone to extract the image feature map from the input image. From this feature map, the RoIAlign layer is used to pool the variable-sized RoIs to a fixed-size region feature map (i.e., $7 \times 7$). In the first branch, this region feature map is fed into two fully connected layers, each having $4096$ neurons, to localize the object location and classify its superclass. In the second branch, this region feature map is fed into three fully connected layers to gradually squeeze the region feature map into a smaller map that fits with the LSTM input. The $LSTM_{roi}$ layer uses the feature from the last fully connected layer to encode the visual information for each RoI, while the $LSTM_{word}$ layer then encodes both the visual features provided by the $LSTM_{roi}$ layer and the input words to generate the object caption.

In practice, we use three fully connected layers with $4096$, $2048$, $512$ neurons, respectively in the second branch. The feature from the last fully connected layer is used as the input for the $LSTM_{roi}$ layer. More formally, the input of the $LSTM_{roi}$ layer is a visual feature vector: $\mathbf{X}^r = (\mathbf{x}_1, \mathbf{x}_2, ..., \mathbf{x}_{n_\ell})$, where $n_\ell$ is the number of LSTM time steps. We note that all the $\mathbf{x}_i \in \mathbf{X}^r$ are identical since they are cloned from one RoI. The $LSTM_{roi}$ layer then encodes its visual input into a list of hidden state vectors $\mathbf{H}^r = (\mathbf{h}_1^r, \mathbf{h}_2^r, ..., \mathbf{h}_{n_\ell}^r)$. Each hidden state vector $\mathbf{h}_i^r$ is combined with one input word to become the input for the second LSTM layer, i.e., the input for the $LSTM_{word}$ layer is a vector: $\mathbf{X}^w = (\mathbf{h}_1^r \oplus \mathbf{w}_1, \mathbf{h}_2^r \oplus \mathbf{w}_2, ..., \mathbf{h}_{n_\ell}^r \oplus \mathbf{w}_{n_\ell})$, where $\oplus$ denotes the concatenation operation and $\mathbf{w}$ is the input word of the object caption. In this way, the network is able to learn the sequential information in the input object caption, while is aware of which object the current caption belongs to (via the concatenation operation).

\textbf{Multi-task Loss} We train the network end-to-end with a multi-task loss $L$ function as follows:

\begin{equation}\label{loss_coarse}
L = {L_{loc}} + {L_{superclass}} + {L_{caption}}
\end{equation}
where $L_{loc}$ and $L_{superclass}$ are defined on the first branch to regress the object location and classify its superclass, $L_{caption}$ is defined on the second branch to generate the object caption. We refer readers to~\cite{Shaoqing2015} for a full description of the $L_{loc}$ and $L_{supereclass}$ loss, while we present the $L_{caption}$ loss in details here.

Let $\mathbf{z}_t$ denote the output of each cell at each time step $t$ in the $LSTM_{word}$ layer. Similar to~\cite{Donahue2014}, this output is then passed through a linear prediction layer $\hat{\mathbf{y}}_t=\mathbf{W}_{z}\mathbf{z}_t+\mathbf{b}_z$, and the predicted distribution $P(\mathbf{y}_t)$ is computed by taking the softmax of $\hat{\mathbf{y}}_t$ as follows:

\begin{equation}
P(\mathbf{y}_t=\mathbf{w}|\mathbf{z}_t)=\frac{\text{exp}(\hat{\mathbf{y}}_{t,w})}{\sum_{w'\in{D}}\text{exp}(\hat{\mathbf{y}}_{t,w'})}
\end{equation}
where $\mathbf{W}_z$ and $\mathbf{b}_z$ are learned parameters, $\mathbf{w}$ is a word in the dictionary $D$. The $L_{caption}$ loss is then computed as follows:

\begin{equation}\label{Eq_loss_caption}
{L_{caption}} = \sum\limits_{i = 0}^{{n_{RoI}}} {\sum\limits_{t = 0}^{{n_\ell}} {P_i({\mathbf{y}_t} = \mathbf{w}|{\mathbf{z}_t})} }
\end{equation}
where $n_{RoI}$ is the number of positive RoIs, $n_\ell$ is the number of LSTM time steps. Intuitively, Equation~\ref{Eq_loss_caption} computes the loss at each word of the current outputted caption for each positive RoI provided by the RNP network.

\textbf{Training and Inference} The network is trained end-to-end for $200k$ iterations using stochastic gradient descent with $0.9$ momentum and $0.0005$ weight decay. The learning rate is empirically set to $0.001$ and keeps unchanging during the training. We select $2000$ RoIs from RPN to compute the multi-task loss. A RoI is considered as positive if it has the IoU with a ground-truth box by at least $50\%$, and negative otherwise. We note that the $L_{box}$ is calculated from both the positive and negative RoIs, while the $L_{supereclass}$ and $L_{caption}$ losses are  computed only from the positive RoIs. In the second branch, each positive RoI is cloned and fed into the first $LSTM_{roi}$ layer, then the word embedding and the hidden states of the first LSTM layer are combined and fed into the second $LSTM_{word}$ layer. This layer converts the inputs into a sequence of outputted words as the predicted object caption. This process is performed sequentially for each word in the predicted caption until the network generates the end-of-caption (\textsf{EOC}) token.

During the inference phase, only the input image is given to the network. We first select top $300$ RoIs produced by RPN as the object candidates. The object detection branch uses these RoIs to assign the object superclass. The results are then suppressed by the non-maximum suppression method~\cite{Girshick2015_deformable_short}. In the captioning branch, all RoIs are also forwarded into two LSTM layers in order to generate the caption for each RoI. The outputted boxes that have the classification score higher than $0.5$ and its associated caption are chosen as the final detected objects. In case there are no boxes satisfying this condition, we select the one with highest classification score as the only detected object.

\subsection{Object Retrieval}\label{Sec_acnn_retrieval}
In this task, rather than generating the object caption, we want to retrieve the desired object in the image given a natural language query. For simplicity, the object is also defined as a rectangle bounding box. The general idea is to select the ``best" (i.e., with the highest probability) bounding box from all region proposals. To this end, our goal is similar to~\cite{guadarrama2014}~\cite{hu2016natural}, however we note that while the authors in~\cite{hu2016natural} focus more on finding the local and global relationship of the query object and other parts of the image, we propose to perform the retrieval task within the concept of the object superclass. In this way, we can train the network end-to-end, while still be able to select the desired object.

\begin{figure}[!htbp] 
    \centering
	\includegraphics[scale=0.34]{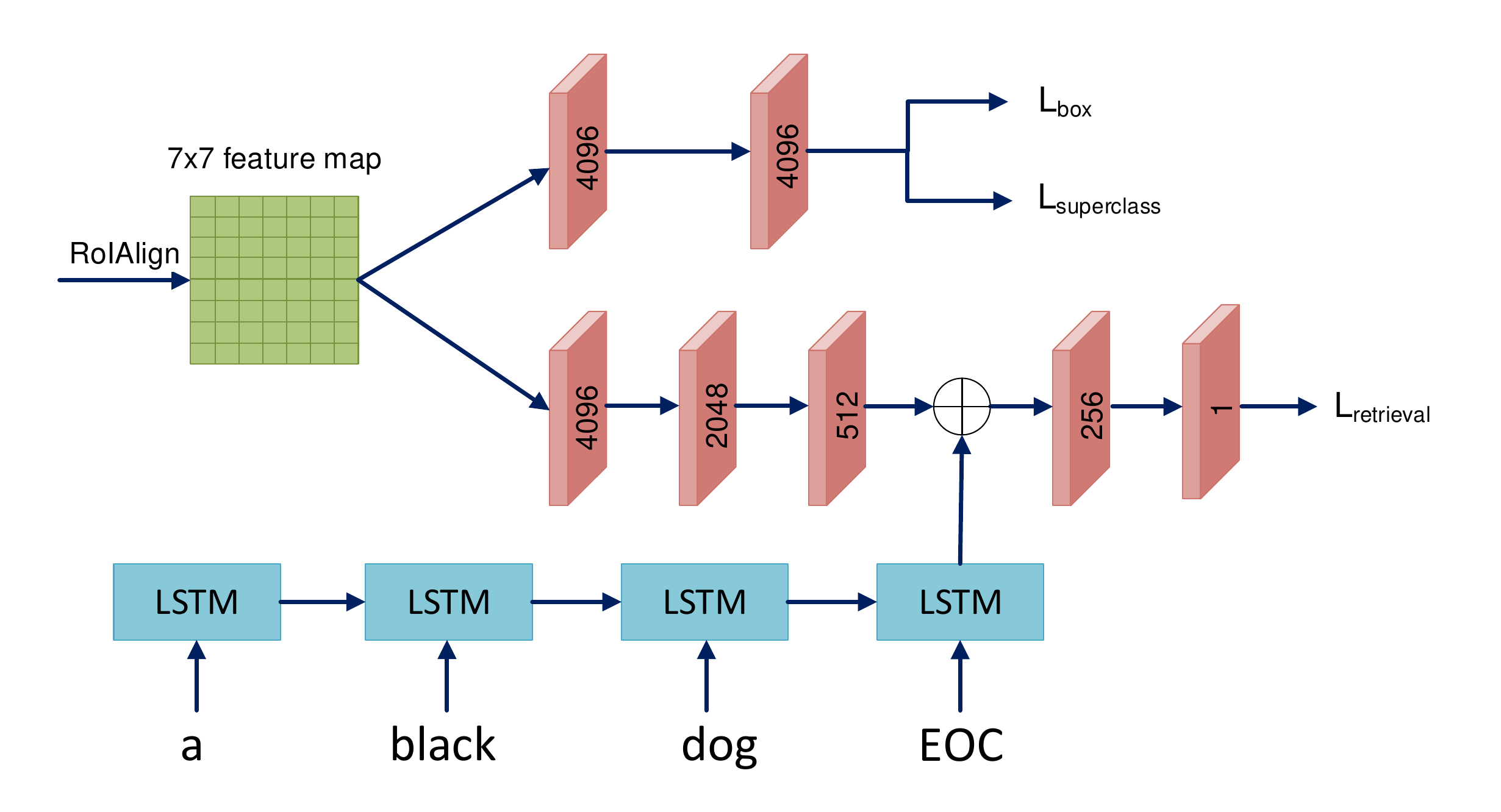} 
    \vspace{1.0ex}
    \caption{An overview of our object retrieval network (ORN). Each input query is encoded by an LSTM layer, then this feature is combined with the RoI feature in order to regress the retrieval score for this RoI.}
    \label{Fig:overview_retrieval} 
\end{figure}

Since we need a system that can generate the object proposals, the RPN network is well suitable for our purpose. We first encode the input query using one LSTM layer. Similar to the object captioning task, we feed each positive RoI into a sequence of fully connected layers (with $4096$, $2048$, $512$ neurons, respectively). The feature map of the last fully connected layer with $512$ neurons is combined with the output of the LSTM layer to create a visual word. This visual word then goes into another fully connected layer with $256$ neurons, then finally the retrieval score is regressed at the last layer with only $1$ neuron. Intuitively, this whole process computes a retrieval score for each positive RoI given the input text query. We note that in parallel with the retrieval branch, the object detection branch is also trained as in the object captioning task. Fig.~\ref{Fig:overview_retrieval} illustrates the details of our object retrieval network.

\textbf{Multi-task Loss} Similar to the object captioning task, we train the network end-to-end with a multi-task loss $L$ function as follows:

\begin{equation}\label{Eq_RetrievalLoss}
L = {L_{loc}} + {L_{superclass}} + {L_{retrieval}}
\end{equation}
where $L_{loc}$ and $L_{superclass}$ loss are identical to the ones in Equation~\ref{Eq_loss_caption}. $L_{retrieval}$ is the sigmoid cross entropy  loss of the retrieval branch and is defined on the positive RoIs as follows:

\begin{equation}
{L_{retrieval}} = \sum\limits_{i = 0}^{{n_{RoI}}} {\log (1 + \text{exp}( - \mathbf{y}_i f({\mathbf{x}_i}))}
\end{equation}
where $\mathbf{y}$ is the groundtruth label (retrieval label) of the current positive RoI, and $f({\mathbf{x}})$ is predicted output of the retrieval branch of the network.

\textbf{Training and Inference} We generally follow the training procedure in the object captioning task to train our object retrieval network. The key difference between these two networks relies on the second branch. In particular, in the object retrieval task, at each iteration, we randomly select one groundtruth object caption as the input query and feed it into the LSTM network. The output of LSTM is then combined with each positive RoI to compute the $L_{retrieval}$ loss for this RoI. Note that, the retrieval groundtruth score of each RoI is automatically reconstructed during training since we know the current input text query belongs to which object (i.e., the positive RoIs associated with the object of the current query has the retrieval score $1$, otherwise $0$).

During the testing phase, the inputs for the network are an image and a text query. Our goal is to select the outputted box with the highest retrieval score. To evaluate the result, each object caption groundtruth of the test image is considered as one input query. This query and the test image are forwarded through the network to generate the object bounding boxes and their retrieval scores. Similar to the object captioning task, we select top $300$ RoIs as the object candidates. These candidates are pruned by the non-maximum suppression process~\cite{Girshick2015_deformable_short}, then the one with highest retrieval score is selected as the desired object. We notice that along with the retrieval score, the network also provides the object superclass and its location via its first branch.

\junk{
We also select top $300$ RoIs as the object candidates. 

by maximizing the log-likelihood of the predicted word.

}

%% file: 4_exp.tex
\section{EXPERIMENTS} \label{Sec:exp}

\subsection{Dataset}

Currently, there are many popular datasets for object detection such as Pascal VOC~\cite{Everingham_PascalVOC} and MS COCO~\cite{MS_COCO}. However, these datasets only provide the bounding boxes groundtruth~\cite{Everingham_PascalVOC} or the caption for the entire image~\cite{MS_COCO}. In the field of referring expressions, we also have the ReferIt dataset~\cite{Kazemzadeh2014} and the G-Ref dataset~\cite{Mao2016}. Although these datasets can be used in the object retrieval task, they focus mostly on the context of the object in the image, while we focus on the fine-grained understanding of each object alone. Motivated by these limitations, we introduce a new dataset (\textit{Flickr5k}) which contains both the object superclass and its descriptions for the fine-grained object understanding. With our new dataset, we can train the network end-to-end to detect the object, generate its caption, or retrieve the object from an input query.

In particular, we select $5000$ images from the Flickr30k dataset~\cite{Plummer_Flick30K}. We only reuse the bounding boxes that come with the Flickr30k dataset then manually assign the superclass and annotate the object captions. Note that, one bounding box only has one specific superclass, while it can have many captions. Totally, our new dataset has $4$ object superclasses (\texttt{people}, \texttt{instruments}, \texttt{animals}, \texttt{vehicles}), $7979$ object bounding boxes, and $18,214$ object captions. The number of bounding boxes for each superclass are $3839$, $475$, $2387$, and $1278$ for the \texttt{people}, \texttt{instruments}, \texttt{animals}, \texttt{vehicles}, respectively. We randomly use $50\%$ of the dataset for training and the remanding $50\%$ for testing. Our new dataset is available at \url{https://goo.gl/MUtyVd}.


\begin{table}[!t]
\centering\ra{1.4}
\caption{Object Captioning Results on Flick5k Dataset}
\renewcommand\tabcolsep{2.5pt}
\label{tb_captioning_result}
\hspace{2ex}

\begin{tabular}{@{}rcccccccc@{}}
\toprule 					 &
\ssmall Bleu\_1  & 
\ssmall Bleu\_2  & 
\ssmall Bleu\_3  &
\ssmall Bleu\_4  & 
\ssmall METEOR  &
\ssmall ROUGE\_L &
\ssmall CIDEr \\

\midrule
OCN1\_VGG16 						&0.564  		& 0.366   		& 0.150				&0.000			& 0.213   		   & 0.559   		& 0.845     \\
OCN1\_ResNet50						&0.566   		& 0.361   		& 0.150				&0.000			& 0.215  		   & 0.563   		& 0.847     \\
OCN1\_ResNet101						&0.571  		& 0.369  		& 0.158				&0.000			& 0.221 		   & 0.570   		& 0.852     \\
\cline{1-8}
OCN2\_VGG16							&0.682   		& 0.555   		& \textbf{0.469}	&\textbf{0.405}	& 0.344   		   & 0.675  		 & 1.628     \\
OCN2\_ResNet50						&0.693   		& 0.568   		& 0.468				&0.403			& \textbf{0.345}   & 0.681  		 & 1.630     \\
OCN2\_ResNet101						&\textbf{0.696} &\textbf{0.570} & \textbf{0.469}	&0.400			& 0.342    		   & \textbf{0.685}   & \textbf{1.632}     \\
\bottomrule
\end{tabular}
\end{table}


\subsection{Implementation}
For both two sub-problems, we use the LSTM network with $512$ hidden units. The number of LSTM timestep $n_{\ell}$ is empirically set to $6$.  Subsequently, the longer captions are cut from the beginning to the sixth word, while the shorter captions are padded with the \textsf{EOC} word until they reach $6$ words. From all the captions, we build a dictionary from the words that appear at least twice, resulting in a dictionary with $866$ words. We use the strategy in~\cite{Kaiming17_MaskRCNN_short}~\cite{AffordanceNet17} to resize the input image to $(600, 1000)$ size. The object proposals are generated by the RPN network with 12 anchors (4 scales and 3 aspect ratios). We use three popular convolutional backbones: VGG16~\cite{SimonyanZ14}, ResNet50 and ResNet101~\cite{He2016} in our experiments. All the networks are trained end-to-end for $200k$ iterations. The training time is approximately $2$ days on an NVIDIA Titan X GPU.

\begin{figure*}[!htbp] 
    \centering
	\includegraphics[width=1.0\linewidth, height=0.5\linewidth]{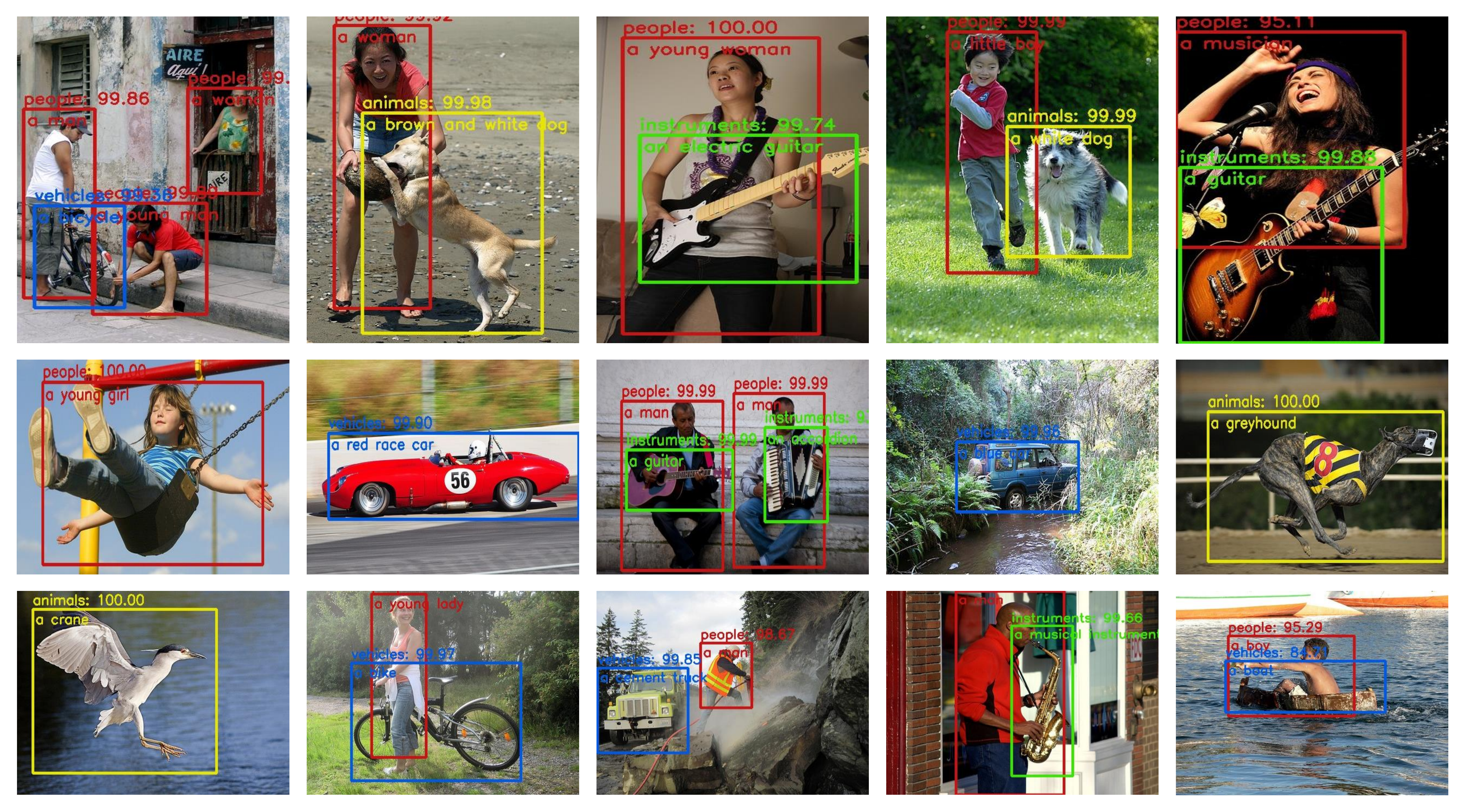} 
    \vspace{0.1ex}
    \caption{Example of prediction results using our OCN2\_ResNet101 network. Besides the correct object detection, the network is able to generate an accurate caption for each object based on its properties (e.g., ``a red race car" vs. ``a blue car", etc.).}
    \label{fig_exp_caption_all} 
\end{figure*}

\subsection{Object Captioning Results}
\textbf{Evaluation Protocol} Although both the traditional object detection and image captioning tasks have the standard metrics to evaluate the results, the object captioning task is more challenging to evaluate since its output contains many information (i.e., the object class, the location, and its caption). In~\cite{Mao2016}, the authors proposed to use human evaluation, however this approach is not scalable. Unlike the system in~\cite{Mao2016} which is not end-to-end and only provides the caption for each bounding box without the superclass information, out network provides all these information. Therefore, we propose to use the standard metrics of image captioning~\cite{Xu16_MSR_Dataset} to evaluate the outputted caption of the bounding boxes that have high classification score. This protocol is also widely used in other problems when the network provides both the detected object and its other information~\cite{AffordanceNet17}~\cite{xiang2017posecnn}.

Table~\ref{tb_captioning_result} summarizes the object captioning results. We compare our object captioning network with two LSTM layers (denotes as OCN2) with the baseline that uses only one LSTM layer (denotes as OCN1, see details in Appendix). Overall, OCN2 clearly outperforms OCN1 by a substantial margin in both three backbones VGG16, ResNet50, and ResNet101. This demonstrates that the way we combine the region feature map with the input caption plays an important role in this task. Intuitively, the approach in OCN2 is more robust than OCN1 since in OCN2 the feature of each positive RoI is combined at every word of the input caption, while in OCN1 the feature is only combined once with the first word. Table~\ref{tb_captioning_result} also shows the ResNet101 backbone achieves the highest performance and outperforms VGG16 backbone. However, this improvement is not very significant. Fig.~\ref{fig_exp_caption_all} shows some example results from our OCN2\_ResNet101 network. It is worth noting that the network is able to generate accurate captions within a superclass based on the object properties (e.g., ``a red race car" vs. ``a blue car", etc.).

\begin{table}[!t]
\centering\ra{1.3}
\caption{Object Retrieval Results on Flick5k Dataset}
\renewcommand\tabcolsep{9.0pt}
\label{tb_retrieval_result}
\hspace{2ex}

\begin{tabular}{@{}rcccccccc@{}}
\toprule 	&
$R@1$  \\

\midrule
SCRC~\cite{hu2016natural}			&68.23\%     \\
ORN\_VGG16 (ours)					&70.61\%     \\
ORN\_ResNet50 (ours)				&74.08\%     \\
ORN\_ResNet101 (ours)				&\textbf{76.36}\%    \\
		
\bottomrule
\end{tabular}
\end{table}

\subsection{Object Retrieval Results}
\begin{figure*}[!htbp] 
    \centering
	\includegraphics[width=1.0\linewidth, height=0.48\linewidth]{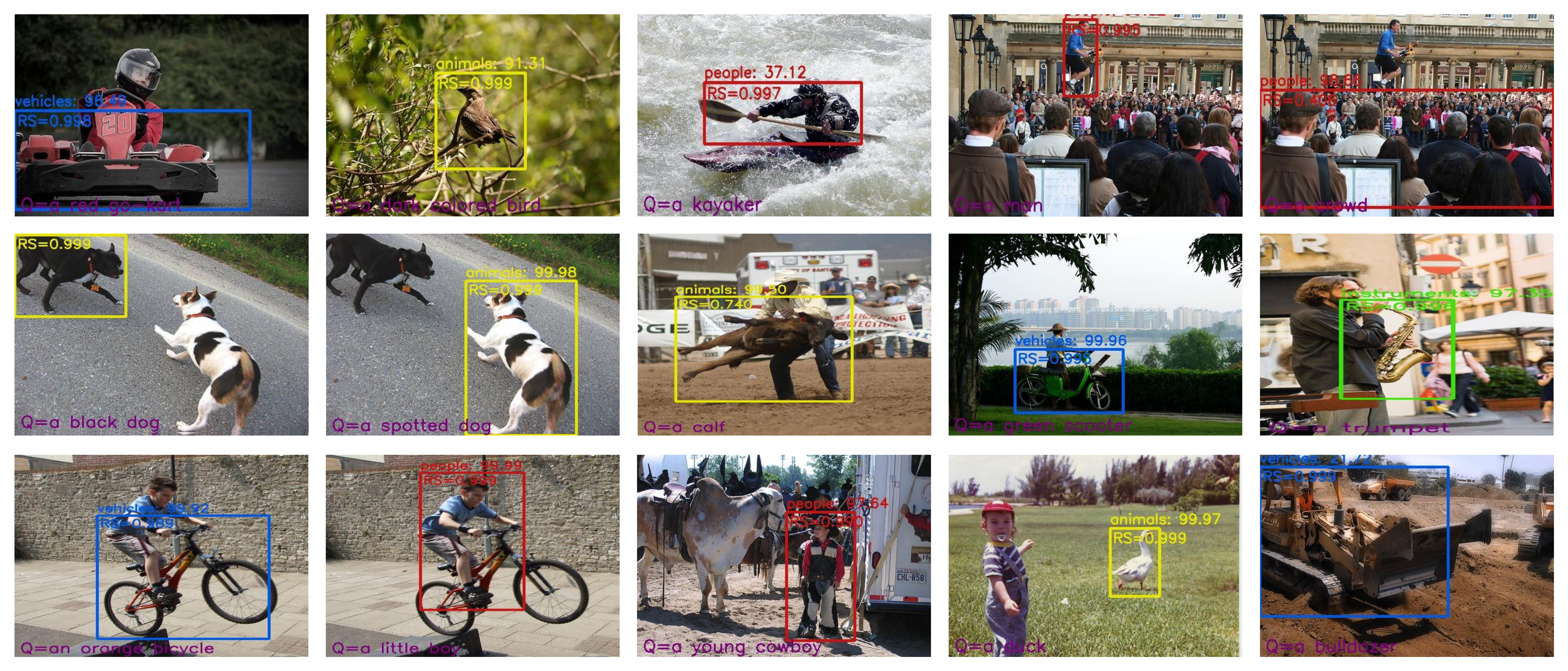} 
    \vspace{0.5ex}
    \caption{Example of retrieval results using our ORN\_ResNet101 network. The network is able to retrieve the correct object in the challenging cases (e.g., ``a black dog" vs. ``a spotted dog"). The text query (Q) is in purple at the bottom of the image. The retrieval score (RS) is denoted inside each detected object. Best viewed in color.}
    \label{fig_exp_retrieval_all} 
\end{figure*}

Similar to~\cite{hu2016natural}, we use the $R@1$ score to evaluate the object retrieval results. The $R@1$ score is the percentage of the predicted box with highest retrieval score being correct. We notice that the predicted box is considered as correct if it has the overlap with the groundtruth box by at least $50\%$ IoU. Table~\ref{tb_retrieval_result} summaries the object captioning results on the Flick5k dataset. Overall, our ORN\_ResNet101 achieves the highest performance with $76.36\%$ of the input queries has the correct retrieval bounding box. This is a significant improvement over SCRC~\cite{hu2016natural}. While we employ an end-to-end architecture to jointly train both the bounding box location and the input query, in SCRC the bounding boxes are pre-extracted and not trained with the network, hence there are many cases the network does not have the reliable bounding box candidates to regress the retrieval score. Furthermore, the inference time of our ORN network is only around $150ms$ per query, which is significantly faster the non end-to-end SCRC approach. Fig.~\ref{fig_exp_retrieval_all} shows some example of retrieval results using our ORN\_ResNet101 network. It is worth noting that the network successfully retrieves the correct object in challenging scenarios such as when there are two dogs (``a black dog" vs. ``a spotted dog") in the image.

\subsection{Ablation Studies}

\textbf{Object Superclass} Unlike the traditional object detection methods~\cite{Shaoqing2015}~\cite{Wei2016} which use the normal object categories (e.g., \texttt{dog}, \texttt{cat}, etc.), we train the detection branch using the superclass (e.g., \texttt{animals}, etc.). With this setup, the object detection branch only provides the location and general knowledge of the object, while the fine-grained understanding of the object is given by the second branch. In the main experiment, we classify all objects into $4$ superclasses in order to keep the basic knowledge of the object categories. However, in applications that do not require the object category understanding, we can consider all the objects belong to one unique superclass (i.e., the \texttt{object} superclass). To this end, we group all the objects of $4$ superclasses into only one \texttt{object} superclass, and then train the captioning and retrieval networks with the ResNet101 backbone as usual.

We follow the same testing procedure as described above. The Bleu\_1, Bleu\_2, Bleu\_3, Bleu\_4, METEOR, ROUGE\_L and CIDEr scores of the OCN2\_ResNet101 network in this experiment are $0.673$, $0.544$, $0.454$, $0.395$, $0.330$, $0.666$, and $1.572$, respectively. While the $R@1$ score of the ORN\_ResNet101 network is $73.06\%$. As we expected, the accuracy of the networks is slightly dropped in comparison with the $4$ superclasses setup, but it is still very reasonable. This demonstrates that the object detection branch can be used to just localize the object location, while the fine-grained knowledge of the object can be effectively learned in the captioning/retrieval branch. More importantly, from this experiment we can conclude that the captioning/retrieval results do not strongly depend on the object classification results of the detection branch, but are actually learned by the captioning/retrieval branch. Compared to the dense captioning framework~\cite{johnson2016densecap} that does not take the object category knowledge into account, or the non end-to-end object retrieval methods~\cite{guadarrama2014}~\cite{hu2016natural}, our approach provides a flexible yet detailed understanding of the object, while still is able to complete both the captioning and retrieval tasks effectively with fast inference time.

\begin{figure}[!htbp]
  \centering
	\subfigure[]{\label{fig:b}\includegraphics[width=0.48\linewidth, height=0.37\linewidth]{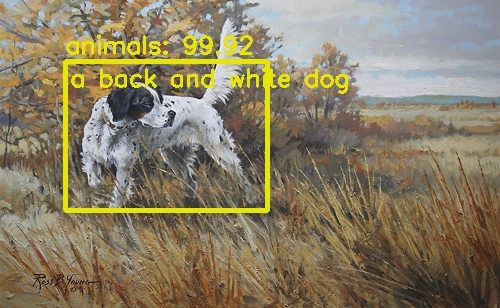}}    
    \subfigure[]{\label{fig:a}\includegraphics[width=0.48\linewidth, height=0.37\linewidth]{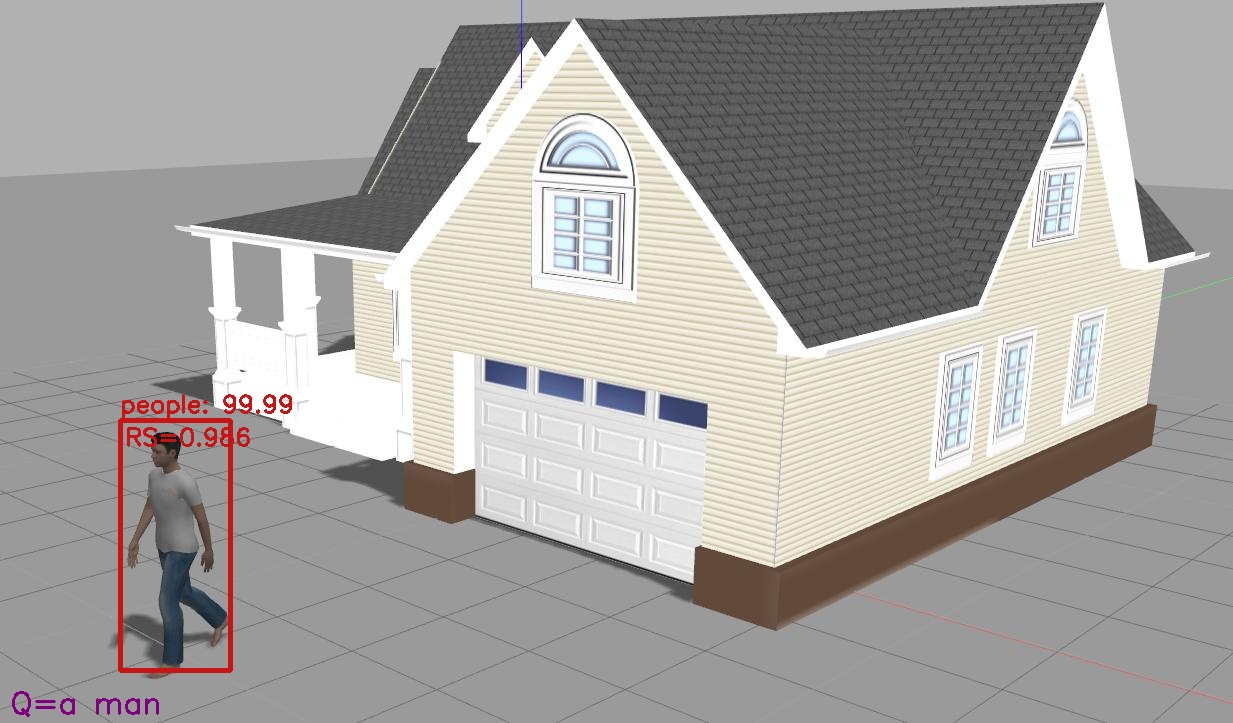}}
 \vspace{2.0ex}
 \caption{Qualitative results. Our networks generalize well under different testing environments. \textbf{(a)} The OCN2\_ResNet101 successfully generates the caption for an object in an artwork image. \textbf{(b)} The ORN\_ResNet101 retrieves the correct object in a simulation image.}
 \label{fig_generalization}
\end{figure}

\textbf{Generalization}
Although we train both of the OCN and ORN networks on a relatively small training set (i.e., there are only $2500$ images in the training set), they still generalize well under challenging testing environments. Fig.~\ref{fig_generalization}-a shows a qualitative result when our OCN2\_ResNet101 network successfully detects the object and generates its caption from an artwork image. In Fig.~\ref{fig_generalization}-b, the ORN\_ResNet101 is able to localize the desired object in an image from Gazebo simulation. Besides the generalization ability, the inference time of both networks is only around $150ms$ per image (or query) on an NVIDIA Titan X GPU, which makes them well suitable for the real-time robotic applications.

\textbf{Failure Cases}
Since we use an end-to-end network to simultaneously train the object detection and the captioning/retrieval branch, the outputted results of the second branch strongly depend on the object location given by the object detection branch. Therefore, a typical failure case in our networks is when the object detection branch outputs the incorrect object location. Fig.~\ref{fig_failure_case}-a and Fig.~\ref{fig_failure_case}-b show two examples when the detection branch misrecognizes the object (i.e., the dog) or is unable to detect the object (i.e., the bird). Similarly, Fig.~\ref{fig_failure_case}-c shows a case when the detection branch is unable to provide the object location for the retrieval branch. We notice that, although in this case the object location is wrong, the retrieval branch is able to assign a very low retrieval score to the wrong object, which shows that it is not confident about the final result.

\junk{

 on the Flick5k dataset

and provide the general knowledge of the object (via its superclass)

This makes our method well suitable for the real-time robotic applications.

 in  information due to the ambiguous visual information , or when the captioning branch generates the incorrect caption due to the ambiguous visual information provided by the object detection branch. In the object retrieval task, the ORN network performs well on the image that has only the current query object, while it becomes challenging when there are many similar objects exist in the input image. Fig.~\ref{fig_failure_case} shows some typical failure cases of our OCN\_ResNet101 network.

discarding the discrimination information of objects within a superclass.\\

To verify if the network is able to learn the discrimination between the objects within a superclass, we conduct the following experiment: From all testing images, we select the ones that have at least two objects belong to the same superclass, then we use the ORN\_ResNet101 network to retrieve the object given its groundtruth query. For each query, we select the outputted bounding box that has highest retrieval score. If this bounding box has the IoU higher than $0.5$ with the groundtruth, we count this as a correct retrieval case.
 
Table~\ref{tb_discrimination} shows the overall probability of the success cases (i.e., number of correct retrieval cases/total queries). We compare this result with the upper-bound baseline of random guess (i.e. $50\%$, when we assume that there are only two objects with the same superclass in all the images. When there are more than two objects with the same superclass, the probability of random guess would be lower than $50\%$). Overall, we achieve the $XX\%$ of success cases in average, which is significantly higher than the upper-bound random guess $50\%$. This demonstrates that the network successfully learn the discrimination between the objects within a superclass. However, we notice that this is still a challenging problem, especially when we have an object that rarely appears during the training.

 }

%% file: 5_conclusions.tex
\section{Conclusions and Future Work}\label{Sec:con}
In this paper, we address the problem of jointly learn vision and language to understand objects in the fine-grained manner. By integrating natural language, we provide a detailed understanding of the object through its caption, while still is able to have the category knowledge from its superclass. Based on the proposed definition, we introduce two deep architectures to tackle two problems: object captioning and object retrieval using natural language. We show that both problems can be effectively solved with the end-to-end hybrid CNN-LSTM networks. The extensive experimental results on our new dataset show that our proposed methods not only achieve the state-of-the-art performance but also generalize well under challenging testing environments and have fast inference time. We plan to release a new large-scale version of our dataset and the full source code of this paper in the future. We hope that these resources can further improve the development of the object captioning and retrieval tasks, making them ready for the real-world robotic applications.

\begin{figure}[!htbp]
  \centering
  \subfigure[]{\label{fig:a}\includegraphics[width=0.32\linewidth, height=0.37\linewidth]{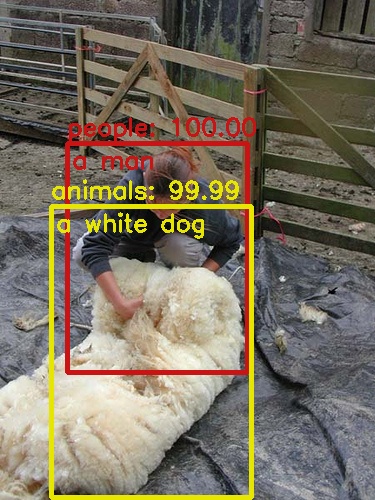}}
  \subfigure[]{\label{fig:b}\includegraphics[width=0.32\linewidth, height=0.37\linewidth]{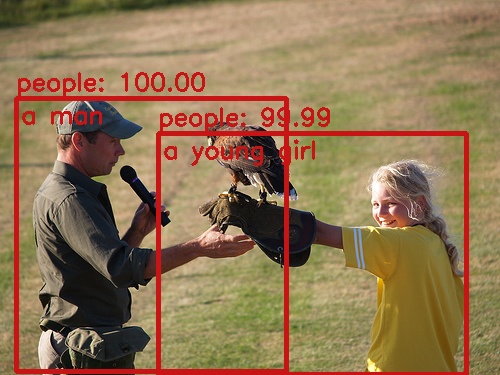}}
    \subfigure[]{\label{fig:b}\includegraphics[width=0.32\linewidth, height=0.37\linewidth]{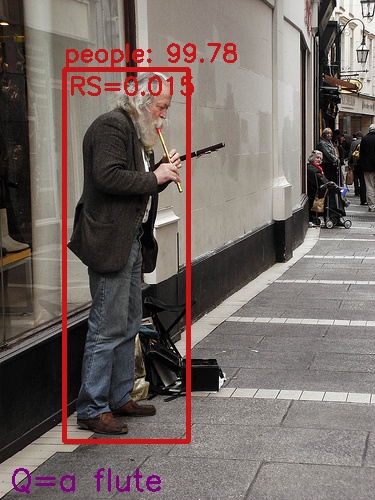}}
     
 \vspace{2.0ex}
 \caption{Some failure cases from the OCN2\_ResNet101 network (\textbf{a}, \textbf{b}), and ORN\_ResNet101 network (\textbf{c}).}
 \label{fig_failure_case}
\end{figure}

\section*{Appendix}
  \label{ocn1_appendix}
The architecture of OCN1 network is as follows:
\begin{figure}[!htbp] 
    \centering
	\includegraphics[scale=0.31]{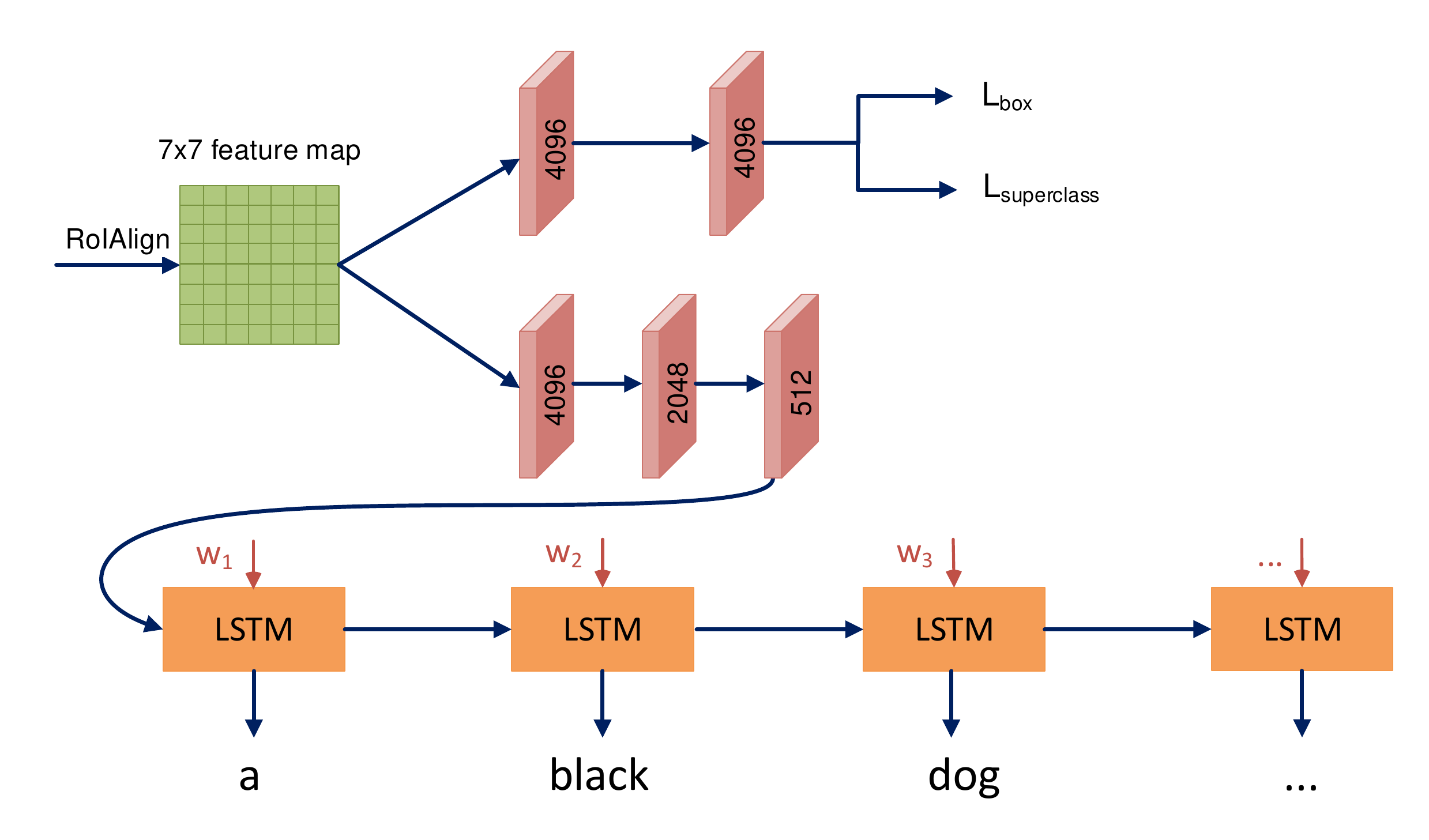} 
    \vspace{1.0ex}
    \caption{An illustration of the OCN1 network with one LSTM layer.}
    \label{Fig:ocn1} 
\end{figure}

While our proposed object captioning network with two LSTM layers (Fig.~\ref{Fig:overview_caption}) combines each input word with the visual feature, the OCN1 network only combines the first word with the visual feature. The experimental results from Table~\ref{tb_captioning_result} show that the OCN1 network has poor performance and cannot effectively generate the long caption.

\section*{Acknowledgment}
\addcontentsline{toc}{section}{Acknowledgment}
Anh Nguyen, Darwin G. Caldwell and Nikos G. Tsagarakis are supported by the European Union Seventh Framework Programme (FP7-ICT-2013-10) under grant agreement no 611832 (WALK-MAN). Thanh-Toan Do and Ian Reid are supported by the Australian Research Council through the Australian Centre for Robotic Vision (CE140100016). Ian Reid is also supported by an ARC Laureate Fellowship (FL130100102). 

\junk
{
 It is also interesting to see how the networks work when we group all the object categories into only one big superclass (i.e., in this case, we train the object detector with only one superclass - the \texttt{object} class). Furthermore, our experiments on the discrimination between objects within a superclass show that this is still a very challenging and unsolved problem. 

Currently, our approach needs two separate networks to detect object affordances. This architecture can't be trained end-to-end as a single network. In future work, we aim to overcome this limitation by developing a new architecture that can detect the object identity and its affordances simultaneously. Another interesting problem is to extend our robotics experiments with more complicated scenarios.
}